\documentclass[runningheads]{llncs}
\usepackage[T1]{fontenc}
\usepackage{graphicx}
\usepackage{booktabs}
\usepackage[misc]{ifsym}

\newcommand{\inlineRoot}[1]{\scalebox{0.8}{$\sqrt{#1}$}}
\newcommand{\corrauthor}{\scalebox{0.7}{\raisebox{1.3ex}{(\Letter)}}}
\usepackage{amssymb}
\usepackage{amsmath}
\usepackage{tabularx}
\usepackage{multirow}
\usepackage{comment}
\usepackage{hyperref}
\hypersetup{
    colorlinks=true,
    linkcolor=blue,
    filecolor=blue,      
    urlcolor=blue,
}
\usepackage{color}
\usepackage{booktabs}
\usepackage{tikz}
\usepackage{microtype}

\makeatletter
\newcommand{\@chapapp}{\relax}%
\makeatother

\usepackage[title,toc,titletoc,header]{appendix}

\usepackage{siunitx}
\newcommand{\appendixsettings}{
    \renewcommand{\thesection}{\appendixname~\Alph{section}}
    \renewcommand{\thetable}{A\arabic{table}}
    \setcounter{table}{0} 
    \renewcommand{\thefigure}{A\arabic{figure}}
    \setcounter{figure}{0}
    \setcounter{page}{1}
}

\usepackage{orcidlink}
\renewcommand{\orcidID}[1]{\orcidlink{#1}}

\begin{document}

\title{Beyond Simulated Benchmarks: Evaluating Motion Representations for Fall Detection Under Real-World Data Scarcity}

\titlerunning{Motion Representations for Fall Detection Under Real-World Data Scarcity}
\author{Timilehin B. Aderinola \inst{1}\corrauthor\orcidID{0000-0002-4770-5871} \and
Ilaria D'Ascanio\inst{3}\orcidID{0009-0000-1568-7149} \and
Luca Palmerini\inst{3}\orcidID{0000-0003-4758-662X} \and
Lorenzo Chiari\inst{3}\orcidID{0000-0002-2318-4370} \and
Jochen Klenk\inst{4}\orcidID{0000-0002-5987-447X} \and
Clemens Becker\inst{4,5}\orcidID{0000-0003-1624-8353} \and
Brian Caulfield\inst{2}\orcidID{0000-0003-0290-9587} \and
Georgiana Ifrim\inst{1}\orcidID{0000-0002-8400-2972}}
\authorrunning{T.B. Aderinola et al.}

%
\institute{School of Computer Science, University College Dublin (UCD), Ireland\\
\and
UCD School of Public Health, Physiotherapy and Sports Science, Ireland
\email{\{timilehin.aderinola,b.caulfield,georgiana.ifrim\}@ucd.ie}\\
\and
Department of Electrical, Electronic and Information Engineering “Guglielmo Marconi”, University of Bologna, Italy
\email{\{luca.palmerini,ilaria.dascanio2,lorenzo.chiari\}@unibo.it}\\
\and
Department of Clinical Gerontology, Robert Bosch Hospital, Stuttgart, Germany\\
\email{\{jochen.klenk,clemens.becker\}@rbk.de}\\
\and
Digital Geriatrics Unit, Heidelberg University Hospital, Heidelberg, Germany
}
\maketitle         

\begin{abstract}
Falls are a major health concern for older adults, and wearable sensors have been widely explored for detecting falls and enabling timely intervention. However, real-world falls are extremely rare: collecting 100 of them requires an estimated 100,000 days of monitoring, resulting in severely limited labelled data for training machine learning models. Consequently, many approaches rely on simulated datasets, often reporting high laboratory performance but limited real-world generalisation. We present a systematic evaluation of motion representations for wearable fall detection under real-world data scarcity. Using accelerometer signals, we compare interval-based, kernel-based, symbolic, and foundation model representations. As an interpretable baseline, we additionally investigate a lightweight symbolic representation that converts short motion segments into symbolic sentences augmented with physically-grounded impact descriptors. Experiments use FallAllD, a simulated falls dataset, and FARSEEING, a clinically verified real-world falls dataset. Through cross-validation, controlled data scarcity, and cross-dataset transfer, we examine how representation choices affect robustness under realistic deployment. Our results reveal that highly parameterised kernel and foundation models excel on simulated data but degrade severely under both data scarcity and domain shift. Although the interval-based representation achieves the strongest absolute real-world performance, augmenting a symbolic representation with physically-grounded impact descriptors yields the smallest degradation under domain shift and retains detection sensitivity under extreme scarcity, albeit at lower precision. These findings highlight the importance of evaluating beyond simulated benchmarks and show that representation choice is critical for deployable fall detection given the scarcity of real-world data.
\end{abstract}

\section{Introduction}
As the proportion of older adults increases worldwide, so does the demand for technologies supporting independent living and remote health monitoring. Among the health risks facing this population, falls are among the most significant, threatening physical safety, long-term independence, and psychological well-being \cite{sucerquia2017sisfall}. Falls are the second leading cause of unintentional injury-related death globally, with an estimated 684,000 fatalities \cite{step_safely_2021} and over 37 million cases requiring medical attention annually \cite{camp2024integrating}. A particular concern is not only the event itself but the ``long lie'': the period during which a person remains on the ground without assistance, which is associated with greater risk of complications and mortality when help is delayed. Fall detection systems have therefore become an important component of geriatric care and remote monitoring \cite{owusu2025litefallnet}. Such systems typically acquire movement during everyday activity, capturing both activities of daily living (ADLs) and falls, followed by signal preprocessing, feature extraction, and classification \cite{liu2023review}.

Many existing fall detection systems emphasise achieving high accuracy using threshold-based heuristics or supervised machine learning models \cite{owusu2025litefallnet}. In controlled laboratory settings, these approaches often report accuracies exceeding 95\%, which can create the impression that fall detection technology is mature and ready for large-scale deployment \cite{nguyen2024model}. However, when evaluated outside controlled laboratory settings, a substantial performance gap often emerges \cite{aderinola2024accurate}. A key reason for this discrepancy is the extreme scarcity of real-world fall data. Because falls are accidental events, collecting sufficient real-world examples is inherently challenging. It has been estimated that capturing 100 real-world falls may require approximately 100,000 days (around 300 person-years) of monitored activity \cite{klenk2016farseeing}. As a result, much of existing research relies on simulated falls performed by young, healthy participants in controlled laboratory environments. While simulated datasets are useful for initial model development, they often fail to reflect the complexity of falls occurring naturally in daily life, leaving a gap between laboratory performance and real-world deployment.

In many fall detection studies, improvements are primarily pursued through increasingly complex model architectures. However, the scarcity of real-world fall data fundamentally limits the ability of such models to learn robust patterns, and high in-domain accuracy offers no guarantee of robustness when models are transferred to real-world conditions. As a result, the choice of motion representation becomes a critical factor: different representations may capture fall dynamics more effectively when only a small number of fall events are available, and may differ in how well they generalise across datasets. Despite this, relatively little work has systematically examined how representation choices influence robustness under data scarcity or their ability to generalise from simulated or private datasets to real-world scenarios.

In this work, using a realistic streaming evaluation pipeline, we systematically evaluate motion representations for wearable fall detection. We compare interval-based, kernel-based, symbolic, and foundation-model representations under realistic constraints of limited real-world fall data and cross-dataset transfer between simulated and real-world datasets, with particular attention to how representations behave when moving from simulated to real-world falls. \textbf{The main contributions of this work are as follows:}

\begin{enumerate}
    \item We present a systematic comparison of interval-based, kernel-based, symbolic, and foundation-model representations for wearable fall detection, using a unified streaming evaluation pipeline with subject-wise splits.
    \item We characterise representation robustness under extreme data scarcity and quantify the simulation-to-reality gap through cross-dataset transfer from simulated falls to clinically verified real-world falls.
    \item As an interpretable case study, we implement FallLM, a lightweight symbolic representation augmenting SAX tokens with impact descriptors, and use its transparency to diagnose failure modes at the token level.
    \item We release our code\footnote{\url{https://github.com/mlgig/fall-lm}} to support reproducible benchmarking of wearable fall detection and related time series event detection tasks.
\end{enumerate}

Our experiments use accelerometer data from the FARSEEING real-world falls dataset \cite{klenk2016farseeing} and the publicly available FallAllD dataset \cite{saleh2020fallalld}, which contains simulated falls performed in controlled laboratory settings. To ensure realistic evaluation, we adopt a streaming event-detection protocol with subject-wise splits. During training, samples are generated using fixed-size overlapping windows extracted from the continuous signals. During testing, signals are processed in a continuous streaming setting, beginning at arbitrary points independent of the fall impact event. For each incoming window, the trained model estimates the probability of a fall, which is then thresholded to produce fall predictions.

The rest of this paper is organised as follows. Section~\ref{sec:related-work} reviews related work on fall detection and time series representation. Section~\ref{sec:methods} describes the datasets, preprocessing steps, and representation methods evaluated in this study. Section~\ref{sec:experiments} presents the experimental setup and evaluation protocols. Section~\ref{sec:results} reports the results and analysis, which are discussed in Section~\ref{sec:discussion}. Finally, Section~\ref{sec:conclusion} concludes the paper and outlines directions for future work.

\section{Related Work}
\label{sec:related-work}

\subsection{Wearable Sensor-Based Fall Detection}
Wearable and ambient sensing technologies have both been explored for automatic fall detection. Ambient devices, such as cameras \cite{bach2026lightweight} or environmental sensors \cite{vignesh2025lidar}, can provide rich contextual information but are limited to fixed monitoring spaces and may raise privacy concerns in sensitive environments. Consequently, wearable inertial measurement units (IMUs), particularly accelerometers, have been widely adopted due to their low cost, portability, and ability to continuously monitor individuals in free-living environments \cite{mohan2024artificial,zhang2024effective}.

Wearable fall detection approaches are typically categorised as threshold-based, machine learning (ML)-based, or hybrid methods \cite{rastogi2021systematic}. Threshold-based techniques \cite{lee2019development} detect falls using predefined rules based on signals such as acceleration magnitude or orientation changes. Although computationally efficient, these methods are limited by high false alarm rates because many activities of daily living (ADLs) exhibit motion patterns similar to falls. Hence, many studies have explored ML-based approaches that involve manual feature extraction from raw motion signals and fall detection with classifiers such as decision trees, support vector machines, and neural networks. More recently, deep learning models \cite{liu2023deep,zafar2025real} have also been applied to automatically learn representations of motion signals. In hybrid systems \cite{palmerini2020accelerometer}, thresholds are often used to pre-filter candidate fall events before applying a learned classifier for final decision making. While a wide range of machine learning models have been proposed for fall detection, relatively little work has systematically examined how different time series representations influence robustness under data scarcity or cross-dataset transfer.

\subsection{Time Series Representations for Wearable Motion Data}
A critical step for fall detection systems is to transform raw sensor signals into compact and discriminative feature spaces that facilitate classification while remaining robust to noise, data leakage and subject variability.
Kernel-based approaches have recently gained popularity for time series classification due to their strong performance and computational efficiency. Methods from the ROCKET family \cite{dempster2020rocket} transform time series using a large number of convolutional kernels. Variants such as MiniRocket \cite{dempster2021minirocket} and MultiRocket \cite{tan2022multirocket} improve efficiency through deterministic kernel selection and optimised pooling operations, while maintaining competitive accuracy on large time series benchmarks.

Interval-based representations summarise signals using statistical features extracted from temporal subsequences. For example, QUANT \cite{dempster2024quant} represents time series through quantiles computed over hierarchical intervals, capturing distributional characteristics of motion signals at different temporal locations.

Symbolic approaches transform real-valued signals into sequences of discrete tokens using segmentation and discretization techniques such as Symbolic Aggregate Approximation (SAX). Methods such as WEASEL \cite{schafer2023weasel} and MrSQM \cite{nguyen2022fast} represent time series as collections of symbolic patterns.

More recently, foundation models for time series have been proposed to learn generalisable representations through large-scale pre-training. Transformer-based models such as Mantis \cite{feofanov2025mantis} use contrastive learning objectives to produce embeddings that can be adapted to downstream tasks such as time series classification.

Despite the diversity of these representation paradigms, many highly accurate methods produce opaque feature spaces that lack clinical interpretability. Furthermore, relatively little work has systematically examined how these representations influence robustness under extreme data scarcity or their ability to generalise from simulated to real-world fall datasets. In this study, we evaluate representative methods from these families, including MiniRocket (kernel-based), QUANT (interval-based), WEASEL (symbolic), MrSQM (symbolic), and the foundation model Mantis. To complement these opaque representations, we additionally include FallLM, a lightweight symbolic representation that encodes motion as sequences of discrete, human-readable tokens augmented with impact descriptors, allowing its decisions to be inspected directly at the token level.

\section{Materials and Methods}
\label{sec:methods}
\subsection{Datasets}

\textbf{FARSEEING} \cite{klenk2016farseeing} is a large collection of clinically verified real-world falls recorded using wearable inertial sensors. It contains 208 falls from 92 participants (mean age $76.1 \pm 12.6$ years). Each recording spans 20 minutes of continuous sensor data with the fall near the midpoint. We use the subset of 150 falls from 54 participants, where the sensors were placed at the fifth lumbar (L5) position and recorded at 100 Hz. The L5 placement provides a stable representation of whole-body motion while reducing noise introduced by limb movements.

\vspace{0.5em}
\noindent
\textbf{FallAllD} \cite{saleh2020fallalld} is a publicly available collection of simulated falls and ADLs performed in controlled settings: 35 fall types and 44 ADLs from 15 participants, recorded at 238 Hz (we resample to 100Hz to match FARSEEING). We use 465 simulated falls from 14 participants with waist-mounted sensors, the placement most comparable to FARSEEING's L5.

\subsection{Data Preprocessing}
\textbf{Signal Aggregation.} The tri-axial acceleration signals are converted into a single acceleration magnitude signal, computed as $M = \inlineRoot{Acc_x^2 + Acc_y^2 + Acc_z^2}$, where $Acc_x$, $Acc_y$, and $Acc_z$ are the anterior–posterior, medial–lateral, and vertical components. Using magnitude reduces sensitivity to sensor orientation and enables consistent processing across devices and placements \cite{gil2023reducing}.

\vspace{0.5em}
\noindent
\textbf{Training Data Segmentation.}
\label{sec:segmentation}
Fall samples are extracted using three-second windows beginning one second before the annotated impact, comprising a one-second pre-impact phase, the impact, and a one-second post-impact phase. This compact window captures the essential dynamics of the fall while remaining suitable for real-time detection. ADL (negative) samples are extracted from fall-free portions using one-second-step sliding windows, retaining only windows whose maximum magnitude exceeds 1.4 g; this removes low-intensity movement while preserving dynamic activities such as walking or turning \cite{palmerini2020accelerometer}. 

\subsection{Streaming Fall Detection Evaluation}
\label{sec:fall-detection}
\begin{figure}[t]
\centering
\includegraphics[width=0.9\textwidth]{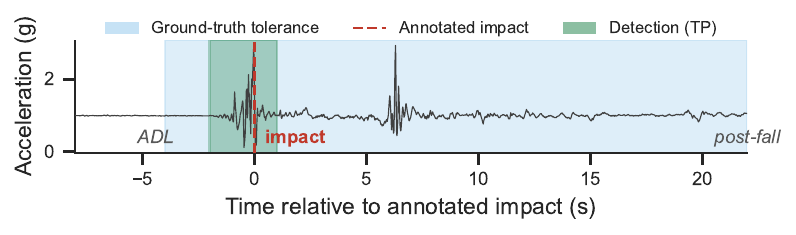}
\caption{A 30-second FARSEEING fall segment showing pre-fall activity, the annotated impact (dashed), and the post-fall lying period. Here, the detection window overlaps the ground-truth tolerance interval (shaded), counting as a true positive. This example is illustrative and not indicative of overall detection precision.} \label{fig:signal-sample}
\end{figure}

Since FARSEEING only provides binary impact labels for each timepoint, we pose fall detection as binary classification (fall vs.\ ADL). To simulate real-world deployment, fall detection is evaluated in a streaming setting where the continuous accelerometer signal is processed sequentially without prior knowledge of fall events, following a recently proposed streaming evaluation protocol \cite{aderinola2025watchyourstep}. The signal is analysed using sliding windows with a one-second step size. For each window, the trained model estimates the probability of a fall event. To account for the temporal context of fall events, detection is evaluated within a tolerance interval surrounding the annotated impact point. This interval includes both the motion leading to the fall and the immediate recovery period following impact. A detection is considered correct if the predicted fall window overlaps with the ground-truth interval (see Fig.~\ref{fig:signal-sample}). Overlap between predicted windows and ground truth is measured using the Intersection over Union, defined as $I\!oU(d,R) = |d \cap R|/|d \cup R|$, where \(d\) denotes the detected window and \(R\) the ground-truth interval. A detection with \(I\!oU(d,R) > 0\) is counted as a true positive, while detections without overlap are considered false positives. If no detection overlaps with the fall interval, the event is counted as a false negative. To avoid registering repeated firings of the sliding window as separate detections, a debounce step retains only the first alarm within a detection episode.

\subsection{Motion Representations}
\label{sec:representations}
Each evaluated representation (Table~\ref{tab:representations}) transforms the acceleration magnitude into a feature space that a classifier then uses. We briefly describe each below.

\begin{table}[t]
\centering
\caption{Representation methods evaluated in this study.}
\label{tab:representations}

\setlength{\tabcolsep}{8pt}
\resizebox{0.85\columnwidth}{!}{
\begin{tabular}{lll}
\toprule
\textbf{Method} & \textbf{Type} & \textbf{Representation}\\
\midrule
WEASEL 2.0 \cite{schafer2023weasel} & Symbolic & Symbolic Fourier Approximation (SFA)\\
MrSQM \cite{nguyen2022fast} & Symbolic & Symbolic Aggregate Approximation (SAX)\\
FallLM & Symbolic & SAX + impact-level tokens\\
MiniRocket \cite{dempster2021minirocket} & Kernel-based & Random convolutional kernels\\
QUANT \cite{dempster2024quant} & Interval-based & Quantile features\\
Mantis \cite{feofanov2025mantis} & Foundation model & Transformer embeddings\\
\bottomrule
\end{tabular}
}
\end{table}

\vspace{0.5em}
\noindent
\textbf{Symbolic Representation.}
Symbolic representations transform continuous signals into sequences of discrete symbols drawn from a finite alphabet. This discretization process reduces sensitivity to noise and allows temporal patterns to be represented as symbolic words that can be analysed using techniques similar to those used in text processing. In this work, we evaluate three symbolic approaches that employ different symbolic transformations and modelling strategies: WEASEL, MrSQM, and FallLM.

\emph{WEASEL 2.0} \cite{schafer2023weasel}, based on the Symbolic Fourier Approximation (SFA), applies a random selection of dilated windows, extracting SFA words at multiple scales and dilations to form a high-dimensional dictionary. The resulting feature vector is classified with a ridge regression classifier.

\emph{MrSQM}
 \cite{nguyen2022fast} discretises segments of the time series into symbolic tokens (SAX or SFA) and constructs features from symbolic subsequences extracted across multiple resolutions, which are then classified using logistic regression.

\begin{figure}[t]
\centering
\includegraphics[width=0.9\textwidth]{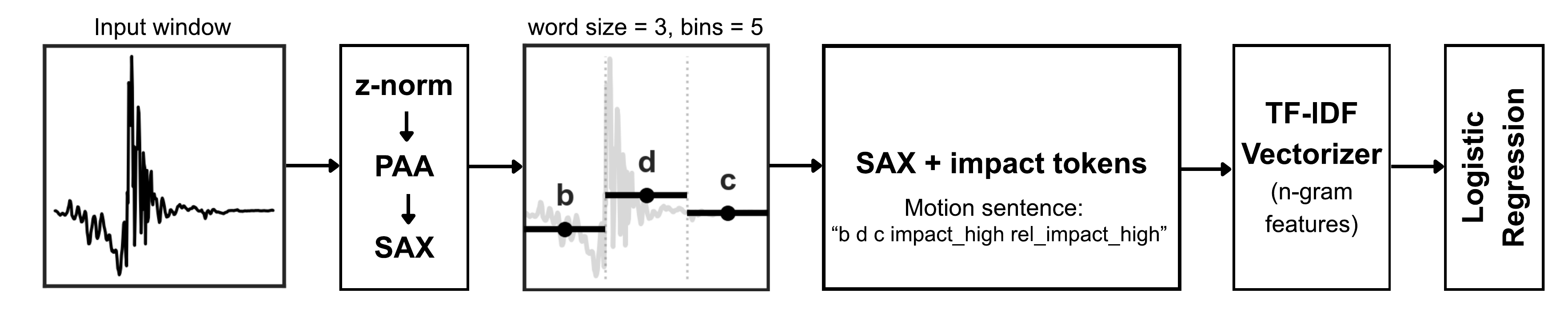}
\caption{Overview of the FallLM pipeline.} \label{fig:FallLM}
\end{figure}

\emph{FallLM.}
We implement \textit{FallLM}, a lightweight symbolic representation for motion signals (see Fig.~\ref{fig:FallLM}). The acceleration magnitude signal is first segmented into short intervals using Piecewise Aggregate Approximation (PAA), after which the aggregated values are discretised into symbolic tokens using SAX. Each window is therefore represented as a short sequence of symbols capturing coarse motion dynamics. SAX discretization uses Gaussian breakpoints derived from the standard normal distribution, yielding a standardised symbolic alphabet.

In addition to the SAX token sequence, we augment each motion sentence with two impact-related tokens derived from the peak acceleration magnitude within the window. The first token encodes the absolute impact level by discretising the peak acceleration into three levels informed by the biomechanical ranges reported in \cite{huynh2021time}: low ($<1.8g$, typical for normal walking), medium ($1.8g$–$2.5g$, associated with deliberate vertical displacement such as climbing stairs), and high ($>2.5g$, associated with hard impacts or falls). Because these thresholds are defined in units of gravitational acceleration ($g$), they remain consistent across datasets and sensor configurations. We show that this coarse, dataset-invariant encoding transfers better than an absent or finer-grained token (see~\ref{app:impact-token}).

To capture contextual information about the impact relative to surrounding motion, we also include a relative impact token that reflects the peak acceleration relative to the magnitude distribution within the window. This token provides a coarse indication of how abrupt the impact is compared to the surrounding movement dynamics. Together, the absolute and relative impact tokens complement the SAX representation by incorporating information about motion intensity while preserving the symbolic structure of the motion sequence. The resulting symbolic sequence is converted into an n-gram bag-of-words representation using a TF-IDF (Term Frequency-Inverse Document Frequency) vectoriser, and a logistic regression classifier is trained on these features for fall detection.

\vspace{0.5em}
\noindent
\textbf{Kernel-Based Representation.} MiniRocket \cite{dempster2021minirocket} transforms the input signal using a large number of fixed convolutional kernels and extracts simple statistics from the resulting feature maps, such as the proportion of positive values. These features form a high-dimensional representation that can be efficiently classified using a linear model (Ridge Regression Classifier).

\vspace{0.5em}
\noindent
\textbf{Interval-Based Representation.} QUANT \cite{dempster2024quant} extracts quantile statistics from hierarchical intervals of the time series, capturing distributional properties of the signal at different temporal locations. These interval features are then used by an ensemble classifier (ExtraTrees Classifier).

\vspace{0.5em}
\noindent
\textbf{Foundation Model Representation.} Mantis \cite{feofanov2025mantis} is a transformer-based model pre-trained using contrastive learning to capture general temporal patterns across datasets. In our experiments, the pre-trained model is finetuned to generate embeddings for fall detection.

\section{Experiments}
\label{sec:experiments}
\subsection{Experimental Setup}
All experiments used Python 3.10 on an Apple M4 Pro (24 GB). Implementations of WEASEL 2.0, MrSQM, MiniRocket, and QUANT were obtained from \texttt{aeon} v0.8.1 \cite{middlehurst2024aeon}, each evaluated as an integrated pipeline with its canonical classifier and default parameters. To address the extreme class imbalance between falls and ADLs, we apply random oversampling of the minority (fall) class within each fold's training split, balancing the classes before model fitting; no classifier-level class weighting is used. For MrSQM, we enable 5 SAX representations and disable SFA, so it serves as a purely SAX-based comparator, directly comparable to FallLM's SAX basis. For Mantis, we implement a scikit-learn-style wrapper around the pretrained model and fine-tune it using PyTorch 2.7.1 with Apple Metal Performance Shaders (MPS) acceleration. Training employs early stopping based on validation loss with a patience of 10 epochs.

For FallLM, the acceleration magnitude signal is discretised using SAX with an alphabet size of 5 and a word size of 3, fixed a priori from the temporal structure of fall events rather than tuned. A word size of 3 maps the 3-second window to the three biomechanical phases of a fall---pre-impact, impact, and post-impact (1 second per symbol)---and keeps motion sentences short enough to remain human-readable for the interpretability analysis (Section~\ref{sec:interpretability}). Each SAX symbol denotes an amplitude level from \texttt{a} (lowest) to \texttt{e} (highest). The default configuration for FallLM, together with sensitivity analyses on the impact tokens and alphabet size, is provided in \ref{app:config}. After appending the two impact tokens, each motion sentence comprises five tokens, vectorised as n-grams (range $(4,5)$) and classified by logistic regression; this n-gram range forces the classifier to weigh the structural shape of the motion together with its impact magnitude.

We conduct three experiments: (1) subject-wise cross-validation on FARSEEING (Section~\ref{sec:cross-validation}), (2) a data-scarcity analysis that progressively reduces the number of training falls (Section~\ref{sec:scarcity}), and (3) a cross-dataset transfer experiment from simulated (FallAllD) to real-world (FARSEEING) falls (Section~\ref{sec:cross-dataset}).

\subsection{Training and Evaluation}
\label{sec:training-and-eval}

\paragraph{Training.}
Each three-second window (300 samples) is labelled with a binary target indicating the presence or absence of a fall event.

\paragraph{Testing.}
The test set consists of unsegmented signals, each containing a single fall annotated with a ground-truth impact index $f$. We define a tolerance interval of $[f-3, f+20)$ seconds around the impact. The 3-second pre-impact allowance captures the falling phase, and the post-impact interval covers the period during which the person may remain on the ground. A detection window overlapping this interval is counted as a true positive (see Section~\ref{sec:fall-detection} for further details).

\paragraph{Evaluation Metrics.}
Given the extreme class imbalance of streaming detection, we report Precision, Recall, and F$_1$ Score rather than accuracy. We additionally report Detection Delay (in seconds) and inference runtime per window as a measure of computational efficiency. Detection Delay is the signed time difference between the detection and the impact index; negative values indicate detection prior to the annotated impact, and smaller absolute values indicate more temporally precise detection.

\section{Results}
\label{sec:results}
\subsection{Cross-validation on Real-World Falls}
\label{sec:cross-validation}
Table~\ref{tab:cv-results} presents the five-fold subject-wise cross-validation results on the FARSEEING dataset (an average of 31 falls per fold). The foundation model, Mantis, obtains the highest F$_1$ score (0.83), with Quant, WEASEL, and MiniRocket close behind. Mantis achieves the highest recall (0.89) and Quant the highest precision (0.84), while WEASEL and MiniRocket maintain a balance between the two. These methods cluster tightly, indicating that with enough labelled real-world falls, interval-based, kernel-based, foundation model, and SFA-based symbolic representations all reach comparable in-domain performance.

The two SAX-based symbolic methods, FallLM and MrSQM, achieve lower F$_1$ scores (0.64 and 0.66). FallLM attains high recall (0.87), but the lowest precision (0.52). In a streaming setting with many ADL windows, this manifests as frequent false alarms, which we examine in Section~\ref{sec:interpretability}. FallLM is the most computationally efficient method (0.01\,ms per sample), though all evaluated methods are already fast enough for real-time and always-on deployment.

Overall, these results show that several representation families achieve strong real-world performance when sufficient labelled data is available, and that the highest in-domain F$_1$ does not single out one representational paradigm. 

\begin{table}[t]
\centering
\caption{Five-fold subject-wise cross-validation results on the FARSEEING dataset. Values are reported as mean(std) across folds.}
\label{tab:cv-results}
\setlength{\tabcolsep}{4pt}
\resizebox{0.85\columnwidth}{!}{
\begin{tabular}{lccccc}
\toprule
\textbf{Model} & \textbf{Runtime (ms)} & \textbf{Precision} & \textbf{Recall} & \textbf{F1} & \textbf{Delay (s)} \\
\midrule
Mantis      & 0.84(0.33) & 0.78(0.05) & \textbf{0.89}(0.07) & \textbf{0.83}(0.02) & -1.28(0.22) \\
Quant       & 0.27(0.06) & \textbf{0.84}(0.08) & 0.80(0.01) & 0.82(0.04) & -0.88(0.15) \\
WEASEL      & 4.77(2.94) & 0.82(0.08) & 0.80(0.05) & 0.81(0.03) & -1.23(0.25) \\
MiniRocket  & 0.27(0.18) & 0.80(0.06) & 0.82(0.08) & 0.80(0.03) & -1.12(0.22) \\
MrSQM       & 3.90(2.49) & 0.64(0.09) & 0.68(0.04) & 0.66(0.05) & \textbf{-0.51}(0.25) \\
FallLM      & \textbf{0.01}(0.00) & 0.52(0.11) & 0.87(0.10) & 0.64(0.09) & -2.31(0.20) \\
\bottomrule
\end{tabular}}
\end{table}

\subsection{Data Scarcity Analysis}
\label{sec:scarcity}
\begin{figure}[t]
\centering
\includegraphics[width=0.75\textwidth]{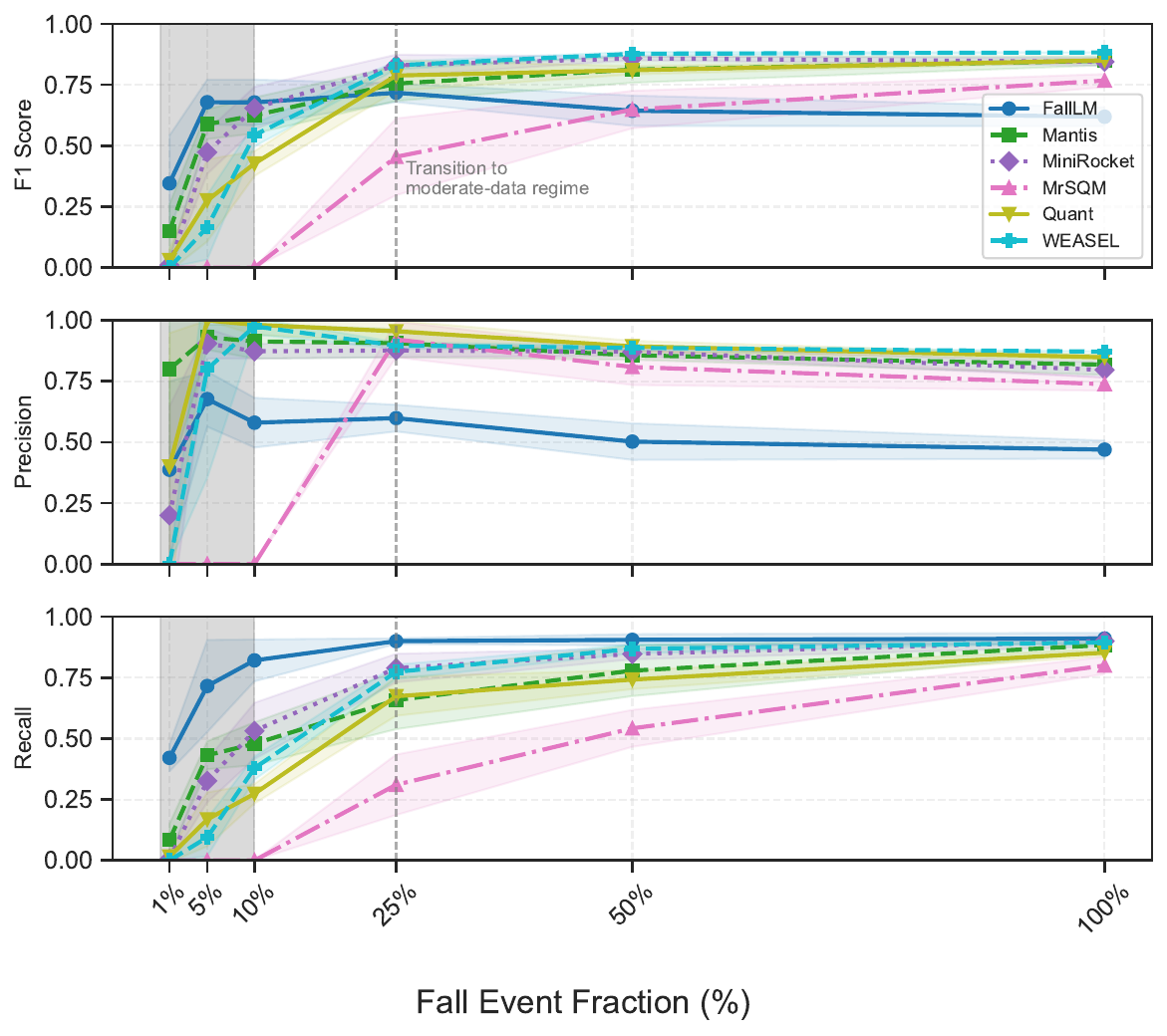}
\caption{Performance under varying levels of real-world fall data availability (FARSEEING). The x-axis represents the fraction of fall events used for training, while all ADL samples are retained. The shaded region highlights the extreme data scarcity regime (1–10\% falls). Results are averaged across five randomly sampled subsets of fall events.} \label{fig:scarcity}
\end{figure}

To examine behaviour under limited real-world data, we train each method on fractions (1\%, 5\%, 10\%, 25\%, 50\%, 100\%) of the 112 fall events in the training set while retaining all negative samples, averaging five runs over randomly sampled fall subsets per fraction. We run this on FARSEEING only, as simulated falls are not scarce. Figure~\ref{fig:scarcity} presents the resulting trends.

With only 1\% of the available fall events (two falls), FallLM is the only method to achieve substantial detection (mean F$_1$ $\approx$ 0.35), recovering falls with minimal supervision but at low and highly variable precision, depending on whether the two sampled falls are representative. Conversely, Mantis fires rarely but almost always correctly (near-perfect precision, low recall, mean F$_1$ $\approx$ 0.11). This conservative behaviour may reflect the priors it inherits from pretraining. QUANT and MiniRocket detect only isolated falls, while MrSQM and WEASEL fail entirely (F$_1=0$). Notably, the methods that perform best with abundant data collapse under extreme scarcity, underscoring that strong in-domain performance does not imply robustness to scarcity.

As more fall events become available, all methods improve rapidly, reaching near-peak performance by roughly 25--50\% of available falls; several show slight precision declines thereafter, likely reflecting the heterogeneity of real falls and their similarity to fall-like ADLs. FallLM, however, plateaus near F$_1$ = 0.64: its recall stays high, but its precision does not improve with additional data, so other methods overtake it on F$_1$ once they accumulate enough falls. Among the symbolic methods, MrSQM does not detect falls until 25\% of falls are available, suggesting that FallLM's scarcity-detection capability stems from its impact tokens rather than from symbolic representation in general.

These results show that the choice of motion representation is decisive under data scarcity. Representations encoding strong, dataset-invariant priors can be particularly valuable when labelled falls are scarce. More broadly, effective fall detection may not require large collections of real-world falls, provided the representation is matched to the scarcity regime.

\subsection{Cross-Dataset Transfer}
\label{sec:cross-dataset}
To assess robustness under domain shift, we train each method on the simulated FallAllD dataset and evaluate it on real-world FARSEEING falls without retraining or adaptation. To enable a direct comparison between in-domain and cross-domain performance, the same FallAllD-trained model is evaluated both on held-out simulated data and on the FARSEEING test folds. Table~\ref{tab:transfer} reports the results as the mean over five subject-wise folds, sorted by relative degradation.

\begin{table}[t]
\centering
\caption{Cross-dataset transfer: models trained on simulated falls (FallAllD) and evaluated on real-world falls (FARSEEING), sorted by relative F$_1$ degradation. Real-world values are mean(std) over five subject-wise FARSEEING folds. P: precision, R: recall.}
\label{tab:transfer}
\setlength{\tabcolsep}{4pt}
\resizebox{0.9\columnwidth}{!}{
\begin{tabular}{ll ccc ccc c}
\toprule
& & \multicolumn{3}{c}{\textbf{Simulated}} & \multicolumn{3}{c}{\textbf{Real}} & \\
\cmidrule(lr){3-5} \cmidrule(lr){6-8}
\textbf{Model} & \textbf{Representation} & F$_1$ & P & R & F$_1$ & P & R & \textbf{F$_1$ Drop\%} \\
\midrule
FallLM     & SAX (symbolic)   & 0.74 & 0.59 & 0.97 & \textbf{0.67}(0.09) & 0.68(0.14) & 0.67(0.11) & \textbf{9.5} \\
Quant      & Interval-based   & 0.95 & 0.97 & 0.94 & 0.62(0.04) & 0.90(0.06) & 0.48(0.06) & 34.7 \\
MrSQM      & SAX (symbolic)   & 0.94 & 0.91 & 0.97 & 0.51(0.08) & 0.42(0.05) & 0.67(0.15) & 45.7 \\
MiniRocket & Kernel-based     & 0.96 & 0.93 & 0.99 & 0.50(0.08) & 0.89(0.10) & 0.36(0.08) & 47.9 \\
Mantis     & Foundation model & 0.98 & 0.97 & 0.99 & 0.50(0.05) & 0.79(0.13) & 0.37(0.06) & 49.0 \\
WEASEL     & SFA (symbolic)   & 0.95 & 0.95 & 0.95 & 0.37(0.09) & 0.85(0.11) & 0.24(0.08) & 62.1 \\
\bottomrule
\end{tabular}}
\end{table}

The simulation--reality gap is substantial and affects all methods. Mantis, MiniRocket, and WEASEL exceed F$_1 = 0.95$ on simulated falls but fall to $0.37$--$0.50$ on real falls, confirming that high simulated performance can badly overestimate real-world effectiveness. QUANT transfers best among the established methods (real F$_1$ 0.62), while FallLM achieves the highest real-world F$_1$ (0.67) and the smallest relative degradation overall (9.5\% drop in F$_1$).

The precision and recall values further reveal that the methods fail in distinct ways. MiniRocket, WEASEL, Mantis, and QUANT retain high precision under transfer but suffer large recall drops, the largest for WEASEL (-0.71). Having learned the morphology of simulated falls, they fire only on the real falls that most resemble them, missing the majority. The two SAX-based methods, FallLM and MrSQM, retain more recall under transfer (drops of 0.30 and 0.31, versus 0.46--0.71 for the others), indicating that the symbolic discretisation helps preserve sensitivity across domains. However, unlike MrSQM, which suffers a precision drop (-0.49), FallLM is the only method whose precision \emph{improves} on real data (from 0.59 to 0.68). We attribute this to the absolute impact magnitude, expressed in physical units of $g$, which remains valid across datasets. This suggests that while symbolic discretisation aids recall retention, it is the physically-grounded impact tokens that additionally preserve precision.

These results show that in-domain accuracy is a poor predictor of cross-dataset robustness, and that the \emph{type} of representational prior matters more than raw performance: representations anchored to dataset-invariant physical quantities transfer better than those that model dataset-specific signal statistics.

\subsection{Symbolic Motion Pattern Analysis}
\label{sec:interpretability}
A key advantage of FallLM's symbolic representation is that the learned model is inspectable. Each feature is a human-readable motion motif, and its logistic regression weight indicates how strongly it pushes a window toward the fall or ADL class (Table~\ref{tab:vocab}). Fall and ADL motifs frequently share amplitude-shape prefixes and differ mainly in their impact token. For example, \texttt{b d c impact\_med} is a strong fall indicator (weight $+1.83$), while \texttt{b d c impact\_low}, identical in shape, pushes towards the ADL class (weight $-0.98$, not among the top motifs shown). Fall motifs are characterised by \texttt{impact\_high} and \texttt{impact\_med} (often with \texttt{rel\_impact\_high}), whereas ADL motifs are dominated by \texttt{impact\_low}. This confirms that FallLM's decisions rest on physically-grounded impact magnitude, and explains its limited in-domain precision: the boundary between falls and ADLs lies in the medium-impact regime, where vigorous activities and genuine falls are least separable. The acceleration waveforms underlying these motifs (see~\ref{app:motif-waveforms}) confirm that the impact tokens correspond to physically meaningful peaks.

\begin{figure}[t]
\centering
\includegraphics[width=0.7\columnwidth]{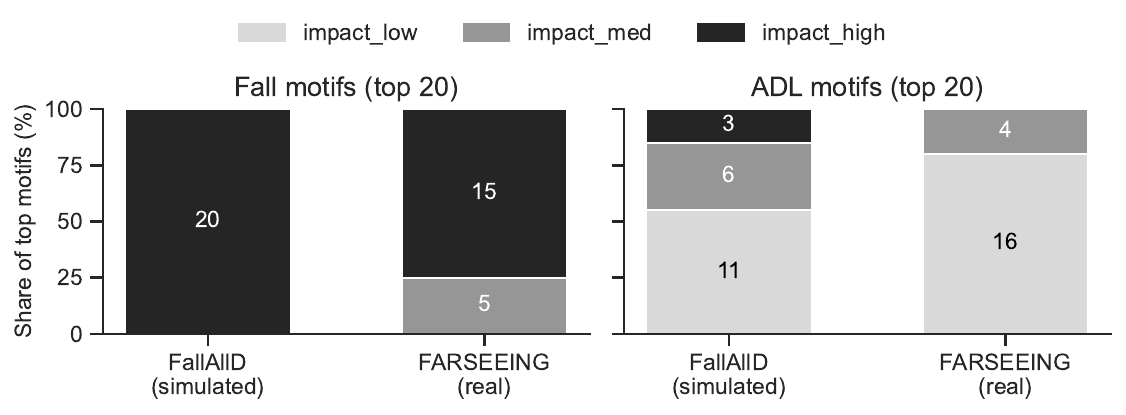}
\caption{Impact-level distribution of the top-20 fall and ADL motifs on FallAllD and FARSEEING. Numbers are motif counts (out of 20).}
\label{fig:vocab-impact}
\end{figure}

Figure~\ref{fig:vocab-impact} further shows the distribution of impact levels among the top fall and ADL motifs learned on each dataset. On FallAllD, every top fall motif carries \texttt{impact\_high}, so the model learns an exclusively high-impact notion of a fall. On FARSEEING, the fall vocabulary extends into the medium-impact range, reflecting that real-world falls span a wider range of intensities. This explains the transfer behaviour of Section~\ref{sec:cross-dataset}: a model trained on FallAllD inherits a high-impact-only fall vocabulary that transfers as a strict, high-precision detector on FARSEEING, where high impact is unambiguous, but which misses lower-impact real falls, capping its recall.

\begin{table}[t]
\centering
\caption{Top fall and ADL motifs learned by FallLM on FARSEEING, by logistic regression weight.}
\label{tab:vocab}
\setlength{\tabcolsep}{6pt}
\resizebox{0.9\columnwidth}{!}{
\begin{tabular}{lr@{\hskip 2em}lr}
\toprule
\multicolumn{2}{c}{\textbf{Fall motifs}} & \multicolumn{2}{c}{\textbf{ADL motifs}} \\
\cmidrule(lr){1-2}\cmidrule(lr){3-4}
Motif & Weight & Motif & Weight \\
\midrule
\texttt{d c impact\_high rel\_impact\_high} & $+2.60$ & \texttt{c c c impact\_low} & $-2.35$ \\
\texttt{b d c impact\_med} & $+1.83$ & \texttt{c c impact\_low rel\_impact\_low} & $-1.76$ \\
\texttt{b d c impact\_high rel\_impact\_high} & $+1.80$ & \texttt{c c impact\_low rel\_impact\_med} & $-1.74$ \\
\texttt{b d c impact\_high} & $+1.80$ & \texttt{c c c impact\_low rel\_impact\_med} & $-1.71$ \\
\texttt{d c impact\_med rel\_impact\_high} & $+1.60$ & \texttt{c c c impact\_low rel\_impact\_low} & $-1.65$ \\
\bottomrule
\end{tabular}}
\end{table}

\section{Discussion}
\label{sec:discussion}
Our results point to a single overarching lesson: in-domain accuracy is a poor predictor of real-world robustness. The representations that dominate the simulated benchmarks degrade severely under domain shift. A practitioner selecting a method on simulated performance alone would choose among the least deployable options. This reframes how wearable fall detection should be evaluated: cross-dataset and scarcity behaviour are not secondary robustness checks but primary selection criteria.

These results suggest that what governs transfer is not model simplicity but the \emph{type} of prior a representation encodes. Methods that learn dataset-specific signal statistics capture the morphology of simulated falls and lose recall when real falls differ in shape; FallLM instead relies on absolute impact magnitude (in units of $g$), a prior whose meaning is invariant across datasets. This is supported by the recall decomposition (Section~\ref{sec:cross-dataset}) and by MrSQM, a second SAX method that, lacking impact anchoring, collapses in precision whereas FallLM's improves.

This robustness has a cost: FallLM is a probe of the principle rather than a deployable detector. Its anchoring makes it imprecise in-domain, where genuine falls and vigorous activities overlap in the medium-impact regime (Section~\ref{sec:interpretability}), and biased toward high-impact events. Trained on simulated falls, whose discriminative signal lies at the high-impact end, it inherits a high-impact-only notion of a fall and systematically misses lower-impact real falls. The same prior that confers robustness thus bounds the method's ceiling: a deployable detector would need to combine impact-anchored robustness with sensitivity to low-impact events.

For practice, these findings carry several implications. Cross-dataset evaluation should be treated as a standard reporting requirement rather than an optional analysis, since simulated performance can greatly overstate real-world effectiveness. Additionally, under data scarcity, representation choice matters more than model capacity.

On the choice of detection paradigm, our binary formulation does not exploit the structure of normal activity that one-class methods model. However, such approaches face a specific difficulty in this domain: high-impact ADLs such as heavy sitting or jumping are statistically anomalous yet are not falls, so detectors keyed on unusual motion would tend to flag them (Section~\ref{sec:interpretability}). A systematic comparison of detection paradigms is left to future work.

Several limitations bound these conclusions. Our analysis rests on a single clinically-verified real-world dataset, and the scarcity of real fall data that motivates this work also limits the breadth of our cross-dataset claims. We evaluate one representative method per family, each with its canonical classifier, so observed differences reflect representation--classifier pipelines rather than representations in isolation. We deliberately study zero-shot transfer to measure the unadapted simulation--reality gap; domain adaptation would likely narrow it but would obscure the quantity we set out to characterise. Finally, the interpretability analysis reflects the highest-weighted motifs of a single model and is illustrative rather than exhaustive.

\section{Conclusion}
\label{sec:conclusion}
We presented a systematic evaluation of motion representations for wearable fall detection under real-world data scarcity and domain shift, comparing interval-based, kernel-based, symbolic, and foundation model approaches on simulated and clinically-verified real-world falls. Across cross-validation, controlled data scarcity, and cross-dataset transfer, we found that the methods strongest on simulated data degrade most under domain shift: the benchmark-topping model is not the most deployable one. Robustness instead tracks the type of prior a representation encodes. Representations anchored to dataset-invariant physical quantities transfer more gracefully than those modelling dataset-specific signal statistics.

As an interpretable case study, FallLM illustrates both sides of this principle: its physically-grounded impact tokens yield the smallest transfer degradation, the highest real-world $F_1$, and detection from as few as two fall examples, but at the cost of in-domain precision and a structural blindness to low-impact falls. As real-world fall data remains extremely scarce, we argue that cross-dataset evaluation should become standard practice, and that reconciling transfer robustness with precision is a key direction for future work. To support reliable benchmarking, we release our code and streaming evaluation pipeline, and continue to work toward the curation of open real-world fall datasets.

\begin{credits}
\subsubsection{\ackname} This study has emanated from research funded by Research Ireland under the Government of Ireland Postdoctoral Fellowship Programme (Grant Number: GOIPD/2025/1434).

\subsubsection{\discintname}
The authors have no competing interests to declare that are
relevant to the content of this article.
\end{credits}
%
%
%
\bibliographystyle{splncs04}
\bibliography{refs}

@article{aderinola2025watchyourstep,
  title={Watch your step: A cost-sensitive framework for accelerometer-based fall detection in real-world streaming scenarios},
  author={Aderinola, Timilehin B and Palmerini, Luca and D'Ascanio, Ilaria and Chiari, Lorenzo and Klenk, Jochen and Becker, Clemens and Caulfield, Brian and Ifrim, Georgiana},
  journal={arXiv preprint arXiv:2509.11789},
  year={2025}
}

@article{gil2023reducing,
  title={Reducing the impact of sensor orientation variability in human activity recognition using a consistent reference system},
  author={Gil-Mart{\'\i}n, Manuel and L{\'o}pez-Iniesta, Javier and Fern{\'a}ndez-Mart{\'\i}nez, Fernando and San-Segundo, Rub{\'e}n},
  journal={Sensors},
  volume={23},
  number={13},
  pages={5845},
  year={2023},
  publisher={MDPI}
}

@article{huynh2021time,
  title={Time-frequency analysis of daily activities for fall detection},
  author={Huynh, Quoc T and Tran, Binh Q},
  journal={Signals},
  volume={2},
  number={1},
  pages={1--12},
  year={2021},
  publisher={MDPI}
}

@inproceedings{nguyen2022fast,
  title={Fast time series classification with random symbolic subsequences},
  author={Nguyen, Thach Le and Ifrim, Georgiana},
  booktitle={International Workshop on Advanced Analytics and Learning on Temporal Data},
  pages={50--65},
  year={2022},
  organization={Springer}
}

@inproceedings{vignesh2025lidar,
  title={LiDAR-based elderly fall detection system for indoor environments using neural network algorithms},
  author={Vignesh, Puthucode Ganesh and Kumar, Nithin and others},
  booktitle={2025 4th International Conference on Sentiment Analysis and Deep Learning (ICSADL)},
  pages={1564--1568},
  year={2025},
  organization={IEEE}
}

@article{bach2026lightweight,
  title={A lightweight and efficient deep learning model for real-time fall detection on edge device},
  author={Bach, Ngo Chi and Van Chi, Nguyen and Cuong, Duong Duc and Tuan, Nguyen Anh and Kieu, Tran Quoc Tu and Phuong, Nguyen Thu and Thien, Nguyen Duy and Thao, Le Quang},
  journal={Biomedical Signal Processing and Control},
  volume={113},
  pages={108942},
  year={2026},
  publisher={Elsevier}
}

@article{zafar2025real,
  title={Real-time activity and fall detection using transformer-based deep learning models for elderly care applications},
  author={Zafar, Raja Omman and Zafar, Farhan},
  journal={BMJ Health \& Care Informatics},
  volume={32},
  number={1},
  pages={e101439},
  year={2025}
}

@article{lee2019development,
  title={Development of an enhanced threshold-based fall detection system using smartphones with built-in accelerometers},
  author={Lee, Jin-Shyan and Tseng, Hsuan-Han},
  journal={IEEE Sensors Journal},
  volume={19},
  number={18},
  pages={8293--8302},
  year={2019},
  publisher={IEEE}
}

@article{zhang2024effective,
  title={An effective deep learning framework for fall detection: model development and study design},
  author={Zhang, Jinxi and Li, Zhen and Liu, Yu and Li, Jian and Qiu, Hualong and Li, Mohan and Hou, Guohui and Zhou, Zhixiong},
  journal={Journal of medical internet research},
  volume={26},
  pages={e56750},
  year={2024},
  publisher={JMIR Publications Toronto, Canada}
}

@article{owusu2025litefallnet,
  title={LiteFallNet: A lightweight deep learning model for efficient real-time fall detection},
  author={Owusu, Emmanuel and Acquah, Isaac and Asare, Michael Asiedu and Yeboah, Benjamin Appiah},
  journal={Digital Health},
  volume={11},
  pages={20552076251386698},
  year={2025},
  publisher={SAGE Publications Sage UK: London, England}
}

@article{rastogi2021systematic,
  title={A systematic review on machine learning for fall detection system},
  author={Rastogi, Shikha and Singh, Jaspreet},
  journal={Computational intelligence},
  volume={37},
  number={2},
  pages={951--974},
  year={2021},
  publisher={Wiley Online Library}
}

@article{step_safely_2021,
  title={Step safely: strategies for preventing and managing falls across the life-course},
  author={{World Health Organization}},
  year={2021},
  publisher={World Health Organization}
}

@article{klenk2016farseeing,
  title={The FARSEEING real-world fall repository: a large-scale collaborative database to collect and share sensor signals from real-world falls},
  author={Klenk, Jochen and Schwickert, Lars and Palmerini, Luca and Mellone, Sabato and Bourke, Alan and Ihlen, Espen AF and Kerse, Ngaire and Hauer, Klaus and Pijnappels, Mirjam and Synofzik, Matthis and others},
  journal={European review of aging and physical activity},
  volume={13},
  pages={1--7},
  year={2016},
  publisher={Springer}
}

@article{palmerini2020accelerometer,
  title={Accelerometer-based fall detection using machine learning: Training and testing on real-world falls},
  author={Palmerini, Luca and Klenk, Jochen and Becker, Clemens and Chiari, Lorenzo},
  journal={Sensors},
  volume={20},
  number={22},
  pages={6479},
  year={2020},
  publisher={MDPI}
}

@article{camp2024integrating,
  title={Integrating fall prevention strategies into EMS services to reduce falls and associated healthcare costs for older adults},
  author={Camp, Kathlene and Murphy, Sara and Pate, Brandon},
  journal={Clinical interventions in aging},
  pages={561--569},
  year={2024},
  publisher={Taylor \& Francis}
}

@article{saleh2020fallalld,
  title={FallAllD: An open dataset of human falls and activities of daily living for classical and deep learning applications},
  author={Saleh, Majd and Abbas, Manuel and Le Jeannes, Regine Bouquin},
  journal={IEEE Sensors Journal},
  volume={21},
  number={2},
  pages={1849--1858},
  year={2020},
  publisher={IEEE}
}

@article{sucerquia2017sisfall,
  title={SisFall: A fall and movement dataset},
  author={Sucerquia, Angela and L{\'o}pez, Jos{\'e} David and Vargas-Bonilla, Jes{\'u}s Francisco},
  journal={Sensors},
  volume={17},
  number={1},
  pages={198},
  year={2017},
  publisher={MDPI}
}

@article{liu2023review,
  title={A review of wearable sensors based fall-related recognition systems},
  author={Liu, Jiawei and Li, Xiaohu and Huang, Shanshan and Chao, Rui and Cao, Zhidong and Wang, Shu and Wang, Aiguo and Liu, Li},
  journal={Engineering Applications of Artificial Intelligence},
  volume={121},
  pages={105993},
  year={2023},
  publisher={Elsevier}
}

@inproceedings{liu2023deep,
  title={Deep Learning-based Fall Detection Algorithm Using Ensemble Model of Coarse-fine CNN and GRU Networks},
  author={Liu, Chien-Pin and Li, Ju-Hsuan and Chu, En-Ping and Hsieh, Chia-Yeh and Liu, Kai-Chun and Chan, Chia-Tai and Tsao, Yu},
  booktitle={2023 IEEE International Symposium on Medical Measurements and Applications (MeMeA)},
  pages={1--5},
  year={2023},
  organization={IEEE}
}

@article{dempster2020rocket,
  title={ROCKET: exceptionally fast and accurate time series classification using random convolutional kernels},
  author={Dempster, Angus and Petitjean, Fran{\c{c}}ois and Webb, Geoffrey I},
  journal={Data Mining and Knowledge Discovery},
  volume={34},
  number={5},
  pages={1454--1495},
  year={2020},
  publisher={Springer}
}

@article{schafer2023weasel,
  title={WEASEL 2.0: a random dilated dictionary transform for fast, accurate and memory constrained time series classification},
  author={Sch{\"a}fer, Patrick and Leser, Ulf},
  journal={Machine Learning},
  volume={112},
  number={12},
  pages={4763--4788},
  year={2023},
  publisher={Springer}
}

@inproceedings{dempster2021minirocket,
  title={Minirocket: A very fast (almost) deterministic transform for time series classification},
  author={Dempster, Angus and Schmidt, Daniel F and Webb, Geoffrey I},
  booktitle={Proceedings of the 27th ACM SIGKDD conference on knowledge discovery \& data mining},
  pages={248--257},
  year={2021}
}

@article{tan2022multirocket,
  title={MultiRocket: multiple pooling operators and transformations for fast and effective time series classification: CW Tan},
  author={Tan, Chang Wei and Dempster, Angus and Bergmeir, Christoph and Webb, Geoffrey I},
  journal={Data Mining and Knowledge Discovery},
  volume={36},
  number={5},
  pages={1623--1646},
  year={2022},
  publisher={Springer}
}

@article{feofanov2025mantis,
  title={Mantis: Lightweight calibrated foundation model for user-friendly time series classification},
  author={Feofanov, Vasilii and Wen, Songkang and Alonso, Marius and Ilbert, Romain and Guo, Hongbo and Tiomoko, Malik and Pan, Lujia and Zhang, Jianfeng and Redko, Ievgen},
  journal={arXiv preprint arXiv:2502.15637},
  year={2025}
}

@article{dempster2024quant,
  title={Quant: A minimalist interval method for time series classification},
  author={Dempster, Angus and Schmidt, Daniel F and Webb, Geoffrey I},
  journal={Data Mining and Knowledge Discovery},
  pages={1--26},
  year={2024},
  publisher={Springer}
}

@article{nguyen2024model,
  title={Model and Empirical Study on Multi-tasking Learning for Human Fall Detection},
  author={Nguyen, Duc-Anh and Pham, Cuong and Argent, Rob and Caulfield, Brian and Le-Khac, Nhien-An},
  journal={Vietnam Journal of Computer Science},
  pages={1--14},
  year={2024},
  publisher={World Scientific}
}

@article{middlehurst2024aeon,
  title={aeon: a Python toolkit for learning from time series},
  author={Middlehurst, Matthew and Ismail-Fawaz, Ali and Guillaume, Antoine and Holder, Christopher and Rubio, David Guijo and Bulatova, Guzal and Tsaprounis, Leonidas and Mentel, Lukasz and Walter, Martin and Sch{\"a}fer, Patrick and others},
  journal={arXiv preprint arXiv:2406.14231},
  year={2024}
}

@inproceedings{aderinola2024accurate,
  title={Accurate and Efficient Real-World Fall Detection Using Time Series Techniques},
  author={Aderinola, Timilehin B and Palmerini, Luca and D’Ascanio, Ilaria and Chiari, Lorenzo and Klenk, Jochen and Becker, Clemens and Caulfield, Brian and Ifrim, Georgiana},
  booktitle={International Workshop on Advanced Analytics and Learning on Temporal Data},
  pages={52--79},
  year={2024},
  organization={Springer}
}

@article{mohan2024artificial,
  title={Artificial intelligence and iot in elderly fall prevention: A review},
  author={Mohan, Deepika and Al-Hamid, Duaa Zuhair and Chong, Peter Han Joo and Sudheera, Kalupahana Liyanage Kushan and Gutierrez, Jairo and Chan, Henry CB and Li, Hui},
  journal={IEEE Sensors Journal},
  volume={24},
  number={4},
  pages={4181--4198},
  year={2024},
  publisher={IEEE}
}

%





\begin{appendices}
\appendixsettings

\section{FallLM Configuration}
\label{app:config}
 
Table~\ref{tab:falllm_config} summarises the fixed FallLM hyperparameters used throughout the paper. These values define the production configuration evaluated in the main paper; the ablations in the following sections vary one parameter at a time while holding all others at these values. In the ablation tables, the rows labelled \emph{default} (\texttt{n\_bins}${=}5$) and \emph{3-level} (impact token) correspond to this configuration.
 
\begin{table}[h]
\centering
\caption{FallLM default configuration. Parameters held fixed across all experiments unless explicitly ablated.}
\label{tab:falllm_config}
\begin{tabular}{ll}
\toprule
Component & Setting \\
\midrule
Window length              & 3\,s (300 samples at 100\,Hz) \\
PAA word size              & 3 (one symbol per 1\,s phase) \\
SAX alphabet size (\texttt{n\_bins}) & 5 (\textit{a}--\textit{e}) \\
Absolute impact token      & 3 levels: low ($<1.8g$), medium ($1.8$--$2.5g$), high ($>2.5g$) \\
Relative impact token      & 3 levels (peak vs.\ within-window distribution) \\
Tokens per motion sentence & 5 (3 SAX + 2 impact) \\
$n$-gram range             & $(4,5)$ \\
Vectoriser                 & TF-IDF \\
Classifier                 & Logistic regression \\
\bottomrule
\end{tabular}
\end{table}

\subsection{Impact-token configuration}
\label{app:impact-token}

We report a transfer ablation (FallAllD~$\rightarrow$~FARSEEING) over three configurations of the impact token, with the SAX alphabet fixed at the 5-bin production default. The \textit{SAX + impact (3-level)} configuration corresponds to FallLM exactly as evaluated in the main paper.

\begin{table}[h]
\centering
\caption{Impact-token configuration ablation. FallAllD columns report the in-domain (simulated) score on a single held-out split; FARSEEING columns report mean (std) across the five FARSEEING transfer folds (real-world). The 3-level configuration is FallLM as used in the main paper.}
\label{tab:ablation_impact_token}
\begin{tabular}{lccc ccc}
\toprule
& \multicolumn{3}{c}{FallAllD (sim)} & \multicolumn{3}{c}{FARSEEING (real, mean (std))} \\
\cmidrule(lr){2-4} \cmidrule(lr){5-7}
Configuration & F$_1$ & P & R & F1 & P & R \\
\midrule
SAX only & 0.670 & 0.808 & 0.573 & 0.418 (0.057) & 0.302 (0.048) & 0.692 (0.108) \\
SAX + impact (3-level) & 0.738 & 0.594 & 0.973 & 0.667 (0.093) & 0.682 (0.136) & 0.668 (0.111) \\
SAX + impact (12-bin)  & 0.794 & 0.684 & 0.945 & 0.464 (0.136) & 0.711 (0.184) & 0.350 (0.119) \\
\bottomrule
\end{tabular}
\end{table}

Table~\ref{tab:ablation_impact_token} compares three impact-token configurations under transfer: no impact token (\textit{SAX only}), the 3-level categorical token used in our final design (\textit{SAX + impact, 3-level}), and a fine-grained 12-bin physical-magnitude token (\textit{SAX + impact, 12-bin}) with bins learned from the training data. With five SAX symbols, shape information alone is already informative in-domain: \textit{SAX only} reaches $F_1 = 0.670$. Adding either impact token improves in-domain performance further, to $F_1 = 0.738$ (3-level) and $F_1 = 0.794$ (12-bin), the 12-bin variant being best in-domain.

The picture reverses under transfer, and the two alternatives fail in opposite ways. The 12-bin variant collapses to $F_1 = 0.464 \pm 0.136$ through a sharp loss of recall ($R = 0.350$): its fine-grained bins, fitted to FallAllD's impact distribution, match too few of FARSEEING's real falls, so it fires rarely (precision $P = 0.711$) but misses most falls. \textit{SAX only} fails in the opposite direction ($F_1 = 0.418 \pm 0.057$): lacking any impact cue, its symbolic vocabulary is not
discriminative enough to suppress false positives, giving low precision ($P = 0.302$) despite reasonable recall. The 3-level categorical token avoids both failure modes, giving by far the best real-world performance ($F_1 = 0.667 \pm 0.093$) and the smallest drop from its in-domain score
($0.738 \rightarrow 0.667$). This confirms that coarse, dataset-invariant magnitude categories generalise better than either no magnitude information or high-resolution
magnitude fitted to a specific population.

\subsection{SAX Alphabet Size}
\label{app:alphabet}

We fix the SAX alphabet size to \texttt{n\_bins}${=}5$, motivated primarily by interpretability. Five amplitude levels yield a compact, human-readable symbol set (\textit{a}--\textit{e}) while retaining sufficient resolution to capture meaningful motion patterns. Smaller alphabets tend to produce highly compressed symbolic vocabularies in which distinct motion patterns collapse to similar motifs, whereas larger alphabets increase symbolic complexity without providing a clear performance advantage. We therefore selected \texttt{n\_bins}${=}5$ as a practical balance between symbolic expressiveness and interpretability.

To verify that the conclusions of the paper are not sensitive to this choice, we sweep \texttt{n\_bins} from 3 to 10 on the in-domain FallAllD dataset (Table~\ref{tab:ablation_alphabet_fallalld}) and, for completeness, report the corresponding sweep on FARSEEING (Table~\ref{tab:ablation_alphabet_farseeing}).

\begin{table}[h]
\centering
\setlength{\tabcolsep}{10pt}
\caption{Alphabet size ablation on FallAllD. Mean (std) over subject-wise 5-fold CV.}
\label{tab:ablation_alphabet_fallalld}
\begin{tabular}{lccc}
\toprule
\texttt{n\_bins} & $F_1$ & Precision & Recall \\
\midrule
3            & 0.745 (0.047) & 0.607 (0.061) & 0.971 (0.028) \\
4            & 0.753 (0.046) & 0.616 (0.061) & 0.973 (0.030) \\
5 (default)  & 0.746 (0.047) & 0.606 (0.062) & 0.976 (0.026) \\
6            & 0.751 (0.048) & 0.615 (0.062) & 0.969 (0.030) \\
7            & 0.727 (0.071) & 0.631 (0.061) & 0.887 (0.173) \\
8            & 0.747 (0.046) & 0.615 (0.059) & 0.956 (0.020) \\
9            & 0.738 (0.064) & 0.666 (0.074) & 0.828 (0.049) \\
10           & 0.741 (0.050) & 0.613 (0.064) & 0.943 (0.036) \\
\bottomrule
\end{tabular}
\end{table}

On FallAllD, performance is essentially flat across the entire range, with $F_1$ varying only between $0.727$ and $0.753$. The default choice of \texttt{n\_bins}${=}5$ is not the highest-scoring configuration, lying marginally below \texttt{n\_bins}${=}4$ ($0.753$), \texttt{n\_bins}${=}6$ ($0.751$), and \texttt{n\_bins}${=}8$ ($0.747$), and well within one standard deviation of all other settings. These results indicate that
alphabet size is not performance-critical in-domain.

\begin{table}[h]
\centering
\setlength{\tabcolsep}{10pt}
\caption{Alphabet size ablation on FARSEEING, reported for completeness. This sweep was \emph{not} used to select \texttt{n\_bins}. Mean (std) over subject-wise 5-fold CV.}
\label{tab:ablation_alphabet_farseeing}
\begin{tabular}{lccc}
\toprule
\texttt{n\_bins} & $F_1$ & Precision & Recall \\
\midrule
3            & 0.645 (0.106) & 0.517 (0.125) & 0.882 (0.083) \\
4            & 0.518 (0.175) & 0.423 (0.225) & 0.775 (0.069) \\
5 (default)  & 0.642 (0.092) & 0.518 (0.106) & 0.866 (0.100) \\
6            & 0.618 (0.142) & 0.514 (0.182) & 0.822 (0.055) \\
7            & 0.637 (0.125) & 0.528 (0.146) & 0.830 (0.075) \\
8            & 0.597 (0.136) & 0.484 (0.171) & 0.826 (0.067) \\
9            & 0.626 (0.145) & 0.505 (0.160) & 0.860 (0.042) \\
10           & 0.633 (0.104) & 0.519 (0.131) & 0.836 (0.070) \\
\bottomrule
\end{tabular}
\end{table}

On FARSEEING, the same broad stability holds for \texttt{n\_bins}~$\geq 5$, with all settings falling within one standard deviation of one another. The only notable deviation occurs at \texttt{n\_bins}${=}4$ ($F_1 = 0.518$), where performance drops sharply relative to neighbouring settings, suggesting that this particular alphabet size merges otherwise distinguishable motion patterns. More generally, FallLM exhibits limited sensitivity to alphabet size across both datasets, indicating that the
representation is robust to moderate changes in symbolic resolution.

Crucially, the alphabet size was not fixed based on FARSEEING performance. Across the entire \texttt{n\_bins}~$\geq 5$ range, FallLM's in-domain behaviour is stable. More broadly, the conclusions of the paper concern differences between representation families rather than a specific FallLM configuration.

\section{Motif Waveforms}
\label{app:motif-waveforms}

\begin{figure}[ht]
    \centering
    \includegraphics[width=1\linewidth]{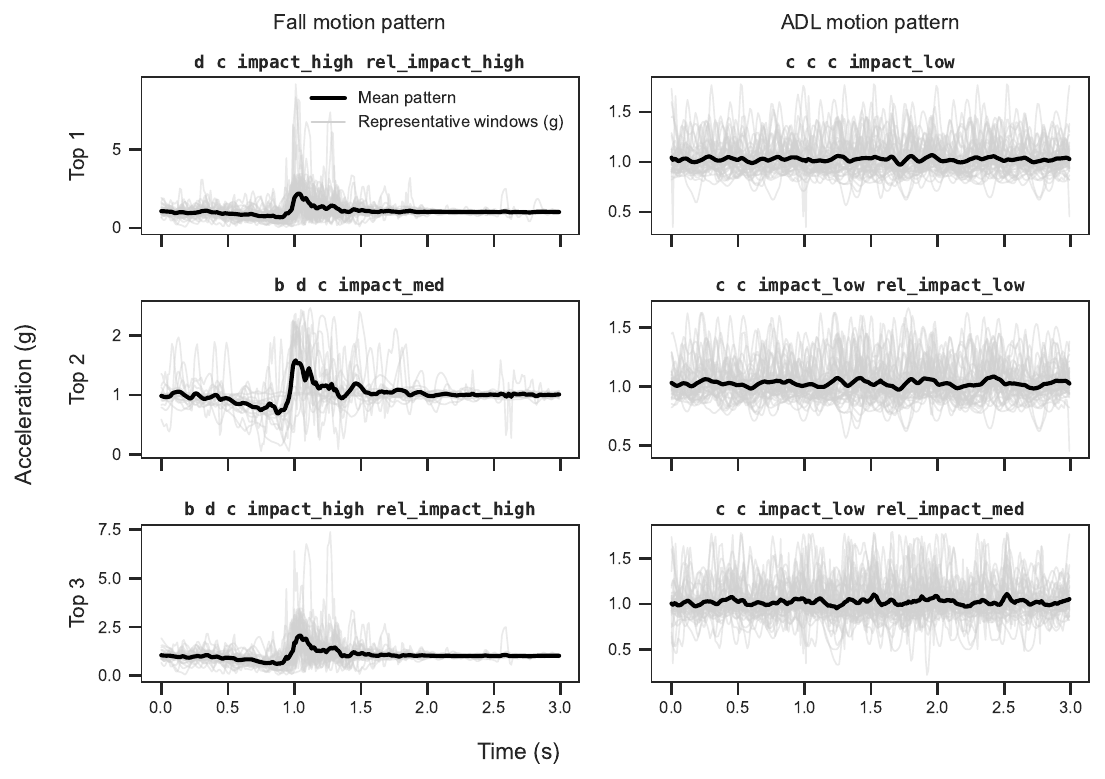}
    \caption{Acceleration waveforms of the three highest-weighted fall (left) and ADL (right) motifs learned by FallLM on FARSEEING. Grey curves show representative windows for each motif; the black curve is their mean.}
\label{fig:motif-waveforms}
\end{figure}

To illustrate that FallLM's symbolic tokens correspond to physically meaningful motion, Figure~\ref{fig:motif-waveforms} shows the raw acceleration waveforms underlying the three highest-weighted fall and ADL motifs learned on FARSEEING (cf.\ Table~\ref{tab:vocab} of the main paper). Each fall motif exhibits a characteristic acceleration peak whose magnitude tracks the absolute impact token: motifs carrying \texttt{impact\_high} peak around 2\,g, while \texttt{impact\_med} motifs peak lower. The ADL motifs, all carrying \texttt{impact\_low}, remain close to 1.5\,g throughout, with no comparable peak. This confirms that the discrete impact tokens are not arbitrary symbols but correspond directly to interpretable acceleration patterns, supporting the token-level analysis in the main paper.

\end{appendices}
\end{document}